\RequirePackage{fix-cm}
\documentclass{svjour3}                     % onecolumn (standard format)
\smartqed  % flush right qed marks, e.g. at end of proof
\usepackage{CJKutf8}
\usepackage{footnote}
\usepackage{lmodern}
\usepackage{mdwlist}
\usepackage{url}
\usepackage[hidelinks,bookmarks=false]{hyperref}
\usepackage[official]{eurosym}
\usepackage{epsfig, times}
\usepackage{algpseudocode}
\usepackage{algcompatible}
\usepackage{longtable}
\usepackage{fancyhdr}
\usepackage{amsmath}
\usepackage{float}
\usepackage{verbatim}
\usepackage{afterpage}
\usepackage{hyperref}
\usepackage{tablefootnote}
\usepackage[toc,acronym,xindy]{glossaries}
\usepackage[colorinlistoftodos,prependcaption,textsize=tiny]{todonotes}

\usepackage{color}
\usepackage{amsfonts}
\usepackage{tabularx}
\usepackage[T1]{fontenc}
\usepackage[utf8]{inputenc}
\usepackage{graphicx}
\usepackage{adjustbox}
\usepackage{subfigure}
\usepackage{arabtex}
\usepackage{utf8}
\usepackage{gensymb}

\usepackage{epsfig}
\usepackage{wrapfig}
\usepackage[ruled]{algorithm2e}

\usepackage{multirow}

\begin{document}

\title{Stylometry Analysis of Multi-authored Documents for Authorship and Author Style Change Detection}
%\subtitle{Disaster event detection, linking and summarization}


\author{\mbox{Muhammad Tayyab Zamir}         \and
        \mbox{Muhammad Asif Ayub}         \and
        \mbox{Asma Gul}         \and  
	    %\mbox{Talhat Khan}         \and
         \mbox{Nasir Ahmad}         \and  
        \mbox{Kashif Ahmad}
       }


\institute{Muhammad Tayyab Zamir \at
              Abasyn University Islamabad Campus, Pakistan \\
              \email{tayyab.zamir999@gmail.com}           %  \\
           \and
           Muhammad Asif Ayub \at
              Department of Computer Systems Engineering, University of Engineering and Technology, Peshawar, Pakistan. \\
              \email{asifayub836@gmail.com}
              \and 
              Asma Gul \at 
              NUST Business School,
National University of Science and Technology, Islamabad, Pakistan. \\ \email{gulasma24@gmail.com}
              \and
                Nasir Ahmad \at 
              University of Western Australia, Australia \\ \email{n.ahmad@uetpeshawar.edu.pk}
              \and
                Kashif Ahmad \at 
              Department of Computer Science, Munster Technological University Cork, Ireland \\ \email{kashif.ahmad@mtu.ie}
              \and
              %A. Al-Fuqaha \at
              %Hamad Bin Khalifa University, Doha, Qatar. \\
              %\email{aalfuqaha@hbku.edu.qa }
}


\maketitle

\begin{abstract}
In recent years, 
the increasing use of Artificial Intelligence (AI)--based text generation tools has posed new challenges in document provenance, authentication, and authorship detection. 
However, %recent developments 
advancements in stylometry %, which involves analyzing an author's writing style, 
%have enabled automatic 
have provided opportunities for 
automatic authorship and author change detection in multi-authored documents using style analysis techniques.
Style analysis can serve as a primary step towards document provenance and authentication through authorship detection. This paper investigates three key tasks of style analysis:  (i) classification of single and multi-authored documents, (ii) single change detection, which involves identifying the point where the author switches, and (iii) multiple author-switching detection in multi-authored documents. We %frame 
formulate all three tasks as classification problems and propose a merit-based fusion framework that integrates several state-of-the-art natural language processing (NLP) algorithms and weight optimization techniques. We also explore the potential of special characters, which are typically removed during pre-processing in NLP applications, on the performance of the proposed methods for these tasks by conducting extensive experiments on both cleaned and raw datasets. Experimental results demonstrate significant improvements over existing solutions for all three tasks on a benchmark dataset. 
\end{abstract}
\keywords{Author Change Detection, Stylometry, Authorship Detection, Author Attribution, Text Classification, LLMs}
\section{Introduction}
\label{sec:introduction}
 In the modern world, the automatic content-generating tools especially the large language models (LLMs) and deep generative AI have made it easier to generate large volumes of grammatically correct and consistent content faster. However, these tools have significantly increased the concerns over the reliability and authenticity of the content in different application domains \cite{pan2023risk,ahmad2022developing}. For instance, in the case of trending news, misinformation or hallucination by such tools may have a significant impact on the policies, trust in governments, and chaos, which may compromise public safety. Similarly, in the education sector, these tools may result in an increase in academic dishonesty in several ways \cite{eke2023chatgpt}.%To overcome these concerns, several efforts have been made by exploring different aspects of visual, audio, and textual content. 
 
 In the case of textual content, one of the key concerns is authorship identification, which mainly involves assessing the authenticity of a document by identifying its original author(s) \cite{kestemont2018overview}. Automatic authorship identification of text documents could prove useful in several application domains such as education (e.g., evaluation and assessment), journalism (e.g., fake news detection and disinformation), law enforcement, and content moderation. Style analysis, which involves exploring and extracting specific patterns in text, is considered a basic step toward developing automatic authorship identification tools. Style analysis enables several interesting tasks and applications allowing the identification of intrinsic features in multi-authored documents. Style change detection (SCD) is one such application of style analysis that could help address growing concerns over document provenance and authentication in various domains \cite{alvi2022style,bevendorff2022overview}.

The existing literature contains several studies that explore different aspects of SCD, which involve tasks such as the classification of single and multi-authored documents, as well as the identification of paragraphs written by the same or different authors (i.e., the locations where the author switches in a multi-authored document). In this work, our focus is on three key tasks of SCD, namely: (i) classification of single and multi-authored documents, (ii) basic style change (i.e., identification of single author-switching in a document), and (iii) real-world style change (i.e., identification of multiple author-switching in a document). These tasks were also part of a benchmark and competition %shared task 
known as PAN-2021\footnote{https://pan.webis.de/index.html}.

Several solutions have been proposed as part of the competition. However, there are still key aspects of the tasks that have not been explored yet. While the existing solutions mostly rely on individual NLP models, we believe that combining these models in a merit-based fusion scheme could significantly improve performance. Moreover, special characters such as punctuation keys, question marks, contractions, and stop and short words could play a vital role in style analysis given that their use varies between authors. Although these special characters are typically removed during pre-processing in NLP applications, it would be interesting to explore their potential in SCD given the nature of the application.

To address these open issues, we propose a fusion-based text classification framework by employing multiple transformers and weight selection/optimization methods. We also analyze the impact of special characters on the three SCD tasks by conducting experiments on both clean and raw datasets. Note that this work is an extension of our previous work \cite{zamir2023document}, which tackled only the first task (i.e., classification of single and multi-authored documents).

The main contributions of this work are summarized as follows.
\begin{itemize}
  \item We propose a novel merit-based late fusion framework that tackles the aforementioned three different tasks of stylometry/style change detection within the text. 
  %These tasks include (i) classification of single and multi-authored documents, (ii) single style change detection, and (iii) style detection at multiple places.
  \item We evaluate the performance of various weight selection/optimization methods including Particle Swarm Optimization (PSO), Nelder-Mead Method, and Powell's method. These methods are employed to assign optimal weights to the individual models based on their performance on the aforementioned tasks. 
  \item Recognizing the significance of special characters, such as punctuation keys, contractions, and short and stop words, in distinguishing writing styles, we conduct extensive experiments on both clean and raw (i.e., unclean) datasets. This exploration aims to assess the impact of utilizing these often-removed special characters during pre-processing on the performance of our proposed methods for style change detection. 
\end{itemize}

The rest of the paper is organized as follows. Section \ref{sec:related_work} provides an overview of the literature. Section \ref{sec:tasks} provides a detailed description of the tasks tackled in this work. Section \ref{sec:methodology} describes the proposed methodology. Section \ref{sec:experiments_setup} provides an overview of the dataset and the experimental setup used in different experiments. Section \ref{sec:results} provides a detailed description of the conducted experiments, experimental results, and key lessons learned. Finally, Section \ref{sec:conclusion} concludes the paper.

\section{Related Work}
\label{sec:related_work}
Stylometry aims to analyze authors' style in written documents. It has been studied extensively in the literature \cite{stamatatos2009survey}. Several interesting works exploring different aspects of the topic \cite{neal2017surveying} have been proposed. Authorship attribution, which relies on SCD techniques, is also one of the key and widely explored aspects of stylometry \cite{zamir2023document}. In this section, we provide a detailed overview of the literature on SCD with a particular focus on the classification of single and multi-authored documents and author change detection.

\subsection{Single Vs. Multi-authored Document Classification}
The classification of single and multi-authored documents is one of the well-explored tasks of SCD. Several aspects of the tasks, such selection of lexical and deep features, have been already explored in the literature \cite{lagutina2019survey}. For instance, Glover et al. \cite{glover1996detecting} analyzed stylometric inconsistencies in multi-authored documents. The authors  explored the associated challenges and potential solutions for the detection of stylometric inconsistencies in collaboratively written documents. They also proposed an SCD method that employs different stylometric features, such as word and sentence length, and distribution of short words for detecting authors' boundaries in multi-authored documents. Similarly, Akiva et al. \cite{akiva2013generic} analyzed and separated segments written by the same authors in multi-authored documents using a clustering technique. Several interesting solutions were also proposed for this task in response to PAN-21 shared task \cite{kestemont2018overview,zangerle2021overview}. For instance, Zhang \cite{zhangstyle} proposed a BERT-based text classification framework for the classification of documents written by single and multiple authors. Strom \cite{strom2021multi} also employed the BERT model along with a classifier trained on lexical features including character-based features (e.g., distinct special characters, spaces, punctuation, etc.,), word-based features (e.g., average word length), and sentence-based features (e.g., Part of Speech (PoS) tags, sentence length, etc.,). Moreover, the authors used words function and contracted words in the lexical feature set. Singh et al. \cite{singh2021writing} also employed lexical features, such as n-grams, POS-Tag, the use of special characters, frequency of words, and the average number of characters per word. A similar set of lexical features is also used by Deibel et al. \cite{deibel2021style}. The feature set includes sentences as well as word features, such as mean sentence length, mean word length, and function word frequency. One of the solutions proposed for the shared task is also based on Siamese Neural Network architecture with bi-directional LSTM trained on lexical features \cite{nath2021style}.

\subsection{Author Change Detection}
Author change detection involves the identification of places where the author switches in a collaboratively written document. There are several compelling studies on this task. For instance, Zhang \cite{zhangstyle} proposed a BERT-based framework that takes pairs of paragraphs from multi-authored documents as input and predicts whether they are written by the same author or not. Nath \cite{nath2021style} proposed a Siamese neural network-based solution where each branch of the Siamese network is fed with a separate paragraph. The author conducted experiments under two experimental setups: one with a bidirectional LSTM model with the Siamese network and the other with a bidirectional GRU (Gated Recurrent Unit) network.

Some studies on author change detection also utilize lexical features. For instance, Deibel et al. \cite{deibel2021style} incorporated a range of lexical features, including mean sentence length, mean word length, and function word frequency at the sentence and work levels, into an LSTM model. Storm \cite{strom2021multi} also employed lexical features in an ensemble method with BERT embeddings. The lexical features include sentence, word, and word-based features, which are then used to train separate classifiers. Finally, the classification scores of the classifiers are combined in an ensemble method. 

In this work, unlike existing solutions, we primarily utilize state-of-the-art transformers,  employing multiple pre-trained transformer-based models both individually and jointly in a merit-based late fusion framework. Additionally, we investigate the potential of special characters, which are typically removed during pre-processing in various NLP applications, on the tasks.

%\section{Problem Statement}
%\label{sec:problem_statement}

%Given a text, determine whether the text is written by a single author or by multiple authors. 
\section{Tasks Description}
\label{sec:tasks}
This paper mainly focuses on style change detection, which involves the identification of the locations in a multi-authored document where the author switches. This task is further divided into three subsequent subtasks: (i) classification of documents written by single and multiple authors, (ii) identification of the position of changes in documents written by multiple authors, and (iii) identification of all the positions where style changes in documents written by multiple authors. These tasks are described in detail below.

\begin{itemize}

\item \textbf{Single vs. Multiple Authors Classification}
The first step in style change detection is to distinguish between documents written by single authors and those written by multiple authors. Multi-authored documents are written by two or more authors and may contain several switches between the authors. We note that the documents are organized into different paragraphs and in multi-authored documents, individual authors are likely to write separate paragraphs as shown in Figure \ref{fig:task1}. Figure \ref{fig:task1}(a) shows an example of a single-authored document, where both paragraphs are written by the same author (i.e., author X). In Figure \ref{fig:task1}(b), the document consists of two paragraphs written by different authors (i.e., author X and author Y). We note that the multi-authored documents must contain at least two paragraphs. Each sample for the task is labeled either '1' or '0' where '0' represents a single author while '1' represents multi-authored documents.

%%%%%%%%%%%%%%%%%%%%%%%%%%%%%%%%%%%%%%%%%
\begin{figure}[!h]
\centering
\includegraphics[width=0.5\textwidth]{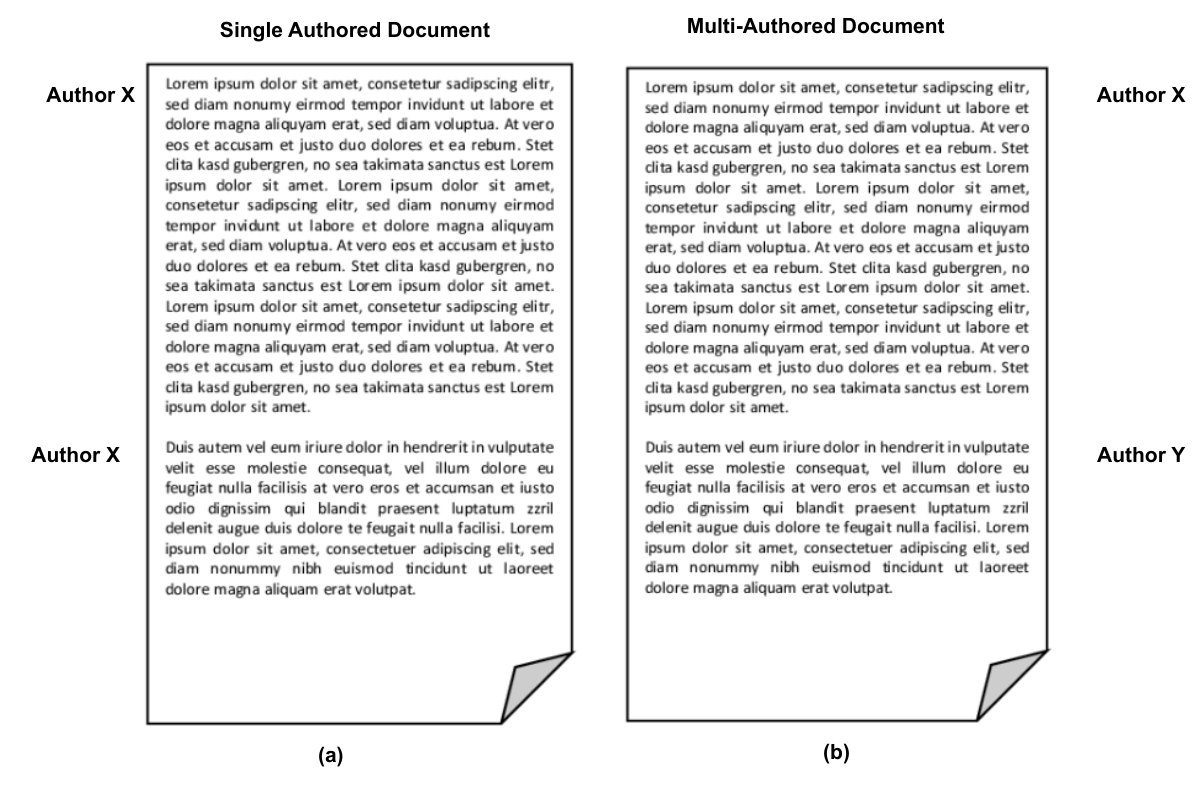}
\caption{Single and multi-authored document samples: (a) single-authored document, (b) multi-authored document \cite{zangerle2021overview}.} 
	\label{fig:task1}
\end{figure}
%%%%%%%%%%%%%%%%%%%%%%%%%%%%%%%%%

\item \textbf{Style Change Basic (Change Positions)}
The second task aims to identify the position where the author switches in a multi-authored document. Similar to task 1, the documents are organized into paragraphs and the proposed solutions predict whether there is a change in author at the boundary of two paragraphs. Figure \ref{fig:task2} displays a sample document consisting of three paragraphs written by two authors (i.e., author X and author Y). The red line highlights the position where the author switches (i.e., from X to Y) between the first two paragraphs, while there is no change in the author at the junction of the second and third paragraphs, both of which are written by the author Y. The annotation for the task is provided in terms of '0' and '1' where '0' represents 'no changes' while '1' indicates that the author has been changed at the position.

%%%%%%%%%%%%%%%%%%%%%%%%%%%%%%%%%%%%%%%%%
\begin{figure}[!h]
\centering
\includegraphics[width=0.5\textwidth]{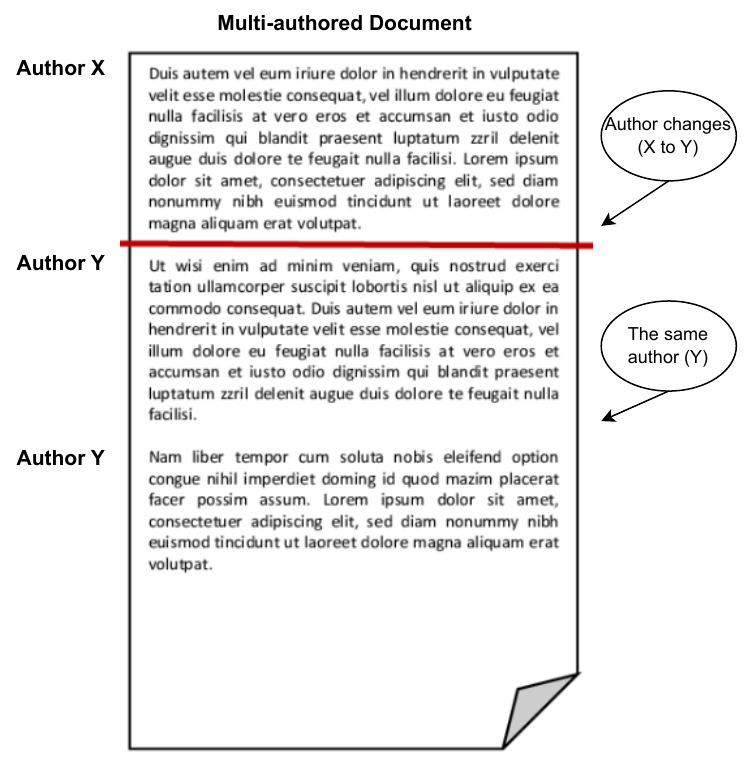}
\caption{A sample multi-authored document with a single author switching \cite{zangerle2021overview}. The red line indicates the position where the author switches from author X to author Y.} 
	\label{fig:task2}
\end{figure}
%%%%%%%%%%%%%%%%%%%%%%%%%%%%%%%%%

\item \textbf{Style Change Real-World (Author Attribution)}
The third task involves the identification of all positions in a document where the writing style changes (i.e., the author switches between the paragraphs). The proposed solutions should be able to automatically assign a paragraph to each author. In simple words, it's an author recognition task where the proposed system must predict the author of each paragraph. Figure \ref{fig:task3} shows an example document from task 3, which is written by three authors, namely author x, y, and author z. The positions where the author switches are represented by a red line. The annotations for this task are provided in terms of $1, 2, 3, ... N$, where $N$ represents the total number of authors contributed to the document.
%%%%%%%%%%%%%%%%%%%%%%%%%%%%%%%%%%%%%%%%%
\begin{figure}[!h]
\centering
\includegraphics[width=0.5\textwidth]{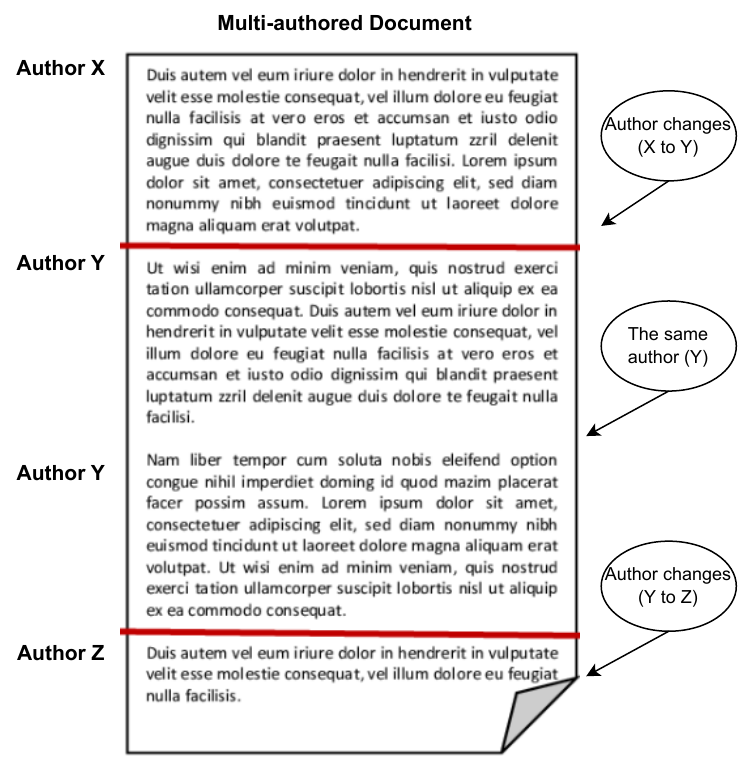}
\caption{A sample multi-authored document with multiple authors switching positions \cite{zangerle2021overview}. The red lines indicate the positions where the author switches (i.e., X to Y and Y to Z) while the absence of the red line between paragraph 2 and 3 indicates that the author remains the same.} 
	\label{fig:task3}
\end{figure}
%%%%%%%%%%%%%%%%%%%%%%%%%%%%%%%%%

\end{itemize}

\section{Methodology}
\label{sec:methodology}
We approach all three tasks as a classification problem and use a similar methodology for all of them. Our approach varies mainly in the pre-processing and data preparation phases, where we adopt customized techniques to prepare the data for classification. 

\begin{figure*}[!h]
\centering
\includegraphics[width=0.85\textwidth]{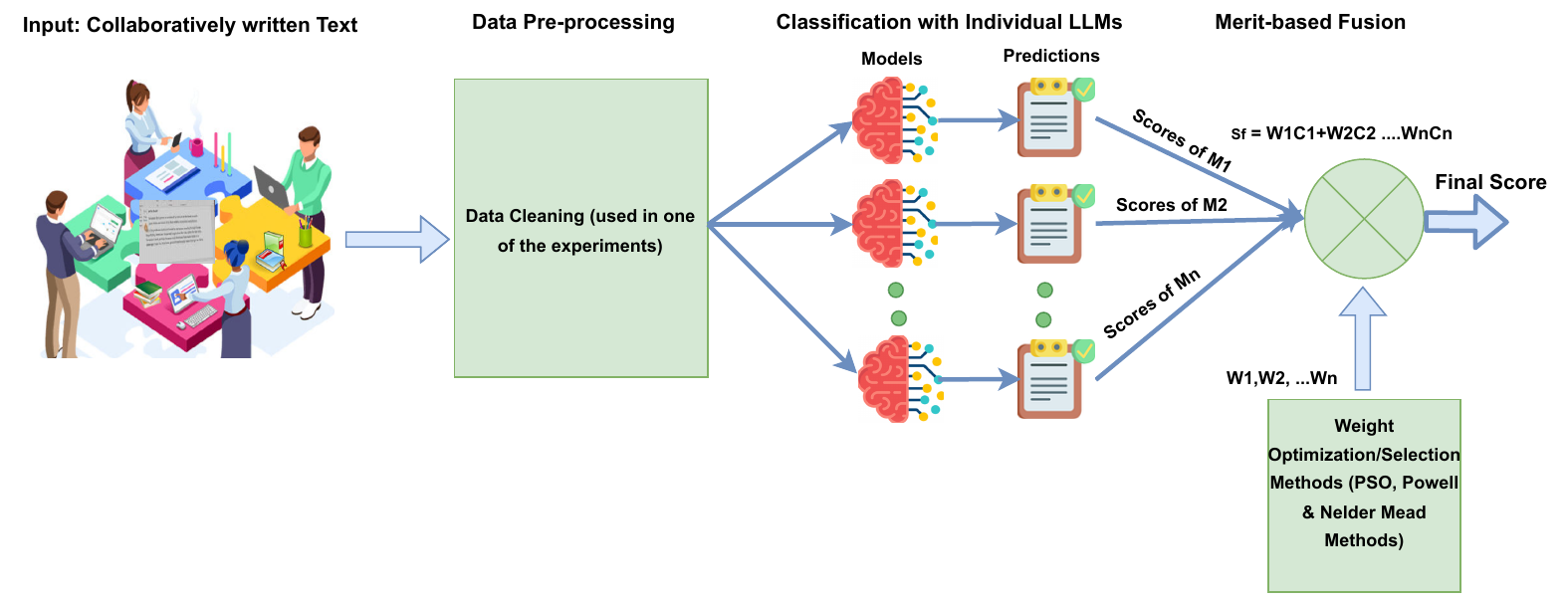}
\caption{The proposed methodology. } 
	\label{fig:methodology}
\end{figure*}

As shown in Figure \ref{fig:methodology}, our methodology is composed of three components: (i) pre-processing, (ii) feature extraction and classification through individual models, and (iii) fusion of the scores obtained from individual models. The last two components are common to all the tasks, and we use the same models and algorithms for classification and fusion. The pre-processing component differs slightly for each task, as we employ techniques that address specific challenges associated with the task and the corresponding data. We describe each of these components in detail below.
%%%%%%%%%%%%%%%%%%%%%%%%%%%%%%%%%%%%%%%%%
%%%%%%%%%%%%%%%%%%%%%%%%%%%%%%%%%
\subsection{Pre-processing}
In the pre-processing step, we mainly focused on data cleaning, which mainly involved the removal of special characters, such as emojis, stop and short words, and the expansion of contractions. We note that the dataset has already been cleaned by the task organizer by removing URLs, code snapshots, and small paragraphs as detailed in Section \ref{sec:experiments_setup}. Based on our previous experience on the task \cite{zamir2023document}, in this work, we extended the scope of pre-processing by exploring different aspects of the tasks. For example, we tried the removal of different characters one by one and analyzed how the performance varies as detailed in Section \ref{sec:results}. Moreover, during our analysis of the data, we observed that in some of the documents, the number of authors and paragraphs was not consistent, which were dropped from the dataset. In addition, we also employed upsampling technique (SMOTE) to balance the dataset. We note that in this work, we did not conduct extensive experiments on the dataset, however, a detailed comparison of results of the proposed methods on tasks with and without data balancing can be found in our earlier work \cite{zamir2023document}. 

\subsection{Feature Extraction and Classification}
For text classification in all the tasks, we rely on various transformer-based language models including BERT (Bidirectional Encoder Representations from Transformers) and its multiple variations, such as DistilBERT, ALBERT, Roberta (Robustly Optimized BERT), and XML-Roberta. The selection of these algorithms is made on the basis of their performances in similar text classification applications, such as natural disaster analysis \cite{said2019natural}, water quality analysis in text \cite{ahmad2022social}, and healthcare \cite{ahmad2022global}. A brief overview of these transformers-based models is provided as follows.

\begin{itemize}
    \item \textbf{BERT}: BERT \cite{devlin2018bert} is one of the pre-trained transformer-based models that have been proven very effective in text analysis. It has been trained on a large collection of Wikipedia unlabeled text and can be fine-tuned for text classification in different application domains. One of the key characteristics of BERT includes its ability to extract contextual information/features through bi-directional learning (i.e., left $\rightarrow$ right and right$\rightarrow$ left).  Despite being very effective in text analysis BERT also has some limitations. For instance, it is computationally intensive. The literature also reports that the model has not been efficiently trained. To overcome these limitations, several variants of BERT have been introduced in the literature, such as RoBERTa and Distil BERT. Moreover, the original model is available in two different configurations having a different number of layers and attention heads. We employed the version of the model with 12 layers and attention heads and 110 million parameters. We used the Tensorflow implementation of the model with Adam optimizer and binary cross entropy loss function. We trained the model for 5 epochs with a learning rate of $1e-5$.
    \item \textbf{DistilBERT}: DistilBERT \cite{sanh2019distilbert} is one of the variants of BERT designed to overcome some limitations of the original implementation. It provides a computationally lighter and faster solution for text analysis. The model is based on the concept of knowledge distillation \cite{hinton2015distilling}, during the pre-trained phase, resulting in a 40\% reduction in model size. While the overall architecture remains similar to the original BERT, DistilBERT removes token-type embeddings and the pooling layer, thereby reducing the number of parameters and model size.  Despite the significant reduction in the size, DistilBERT retains 97\% of the knowledge from the original model. Similar to BERT-base, we used the  Tensorflow implementation,  employing binary cross-entropy, the Adam optimizer, a learning rate of 1e-5, and training for 5 epochs. 

    \item \textbf{RoBERTa}: RoBERTa \cite{liu2019roberta} is another transformer model that emerged as a result of the efforts for improving the BERT model. RoBERTa is built on top of the BERT by making several changes to the training process.  For instance, a larger mini-batch size and a higher learning rate have been used in training RoBERTa. Similarly, the model has been trained on a larger dataset for a longer time. More importantly, the BERT's next sentence pre-training objective has also been removed from RoBERTa.  
     \item \textbf{XML-RoBERTa}: XLM-RoBERTa is a variant of RoBERTa that is mainly used for multi-lingual text analysis. It is trained on 2.5TB of data containing text from 100 different languages. Though all the text samples in our dataset are in English, XML-RoBERTa could complement the other models in the fusion as demonstrated in similar kinds of applications \cite{ahmad2022social}. We used the same parameter and hyper-parameter settings for RoBERTa (i.e., learning rate = $1e-5$, number epochs = 5, loss function = binary-cross entropy, and Adam optimizer). 
      \item \textbf{ALBERT}: ALBERT \cite{lan2019albert} is another variant of BERT introduced to minimize the size of the original BERT model. To this aim, two different parameter reduction techniques including (i) parameter sharing and (ii) a reduced size of embedding, have been employed resulting in a significant reduction in memory requirements as well as training speed. Moreover, its modeling inter-sentence coherence capabilities also make it a preferred choice for multi-sentence input text. In this work, we used cross-entropy as a loss function with an Adam optimizer using a learning rate of 1e-5 for 5 epochs. 
   % \item \textbf{LSTM}: LSTM is a special type of Recurrent Neural network (RNNs) and is able to learn long-term dependencies. The ability to learn long-term dependencies makes them a suitable choice in a diversified list of applications, such as speech translation. LSTMs have also been widely used in text processing applications \cite{leevy2020short}. Similar to all RNNs, LSTMs also possess chain-like structures, however, in contrast to other RNNs, the repeating module of LSTMs has a different structure by having four neural network layers instead of a single one. 
  %  \item \textbf{biLSTM}: Bidirectional LSTM (biLSTM) is one of the key variants of LSTM. In contrast to standard LSTMs, input flows in both directions in biLSTMs. In other words, we can say biLSTMs are composed of two LSTMs where one takes the input in a forward and the other one takes the input in a backward direction. Such a structure makes it a suitable choice for NLP tasks. 
\end{itemize}
For a fair comparison, we use the same hyperparameters for all the models, which are summarized in Table \ref{tab:hyperparameters}.

%%%%%%%%%%%%%%%%%%%%%%%%%%
\begin{table}[!h]
\caption{A summary of the hyper-parameters used in the experiments.}
		\label{tab:hyperparameters} 
\begin{tabular}{|c|c|}
\hline
\textbf{hyper-parameter} & \textbf{Value} \\ \hline
	Learning Rate & 1e-5  \\ \hline
        Epochs & 5 \\ \hline
        mini-batch size &  32\\ \hline
        Dropout &  0.1\\ \hline
	Optimizer/Training Solver & Adam \\ \hline
	Cost Function & binary cross-entropy \\ \hline
\end{tabular}
\end{table}
%%%%%%%%%%%%%%%%%%%%%%%%%%%%%%%%%%
\subsection{Fusion}
For fusion, we mainly rely on merit-based late fusion where multiple weight selection and optimization methods are employed to combine the classification scores of the individual models using Equation \ref{eqn:fusion}.

In the equation, $F(0,1)$ represents the combined classification score while  $w_{n}$ and $p_{n}$ denote the weight to be assigned to the model and the classification score obtained with the individual (i.e., nth) model, respectively. The weight optimization methods are used to calculate the values of the weights assigned to each model (i.e., $w_{1},w_{2},w_{3}, ...w_{n} $). 
%%%%%%%%%%%%%%%%%%%%
\begin{equation}
\label{eqn:fusion}
F(o,1)=w_{1}P_{1}(0,1)+w_{2}P_{2}(o,1)+w_{3}P_{3}(0,1)+....+w_{n}P_{n}(0,1)
\end{equation}
%%%%%%%%%%%%%%%%%%%%%%%%

%%%%%%%%%%%% Algorithm %%%%%%%%%%%%
\begin{algorithm}[htbp]
\footnotesize
\caption{Proposed Merit-based Fusion}
\label{fusion_algo}
\begin{algorithmic}[1]
\STATEx \textbf{Input}: $N$: number of models, $w_{1},w_{2},w_{3}, ...w_{n}$: weights obtained through the weight selection/optimization method, $p_{1}(0,1),p_{2}(0,1),p_{3}(0,1), ...p_{n}(0,1)$: classification posterior probabilities obtained by the individual models for both classes 0 and 1, $M$: number of samples in the dataset.
\STATEx \textbf{Output}: $F(0,1)$: combined classification scores, and $L_{i}$: final label of the sample i.

\STATEx
%\begin{center}
%\textbf{\textit{// Server initialization}}
%\end{center}
\FOR {$i$ = 1 to $M$}
\FOR {$j$ = 1 to $N$}
%\STATE Pick random model $M_c$ for cluster $c$
%\STATE Initialize the global model $M_c$
%\STATE Pick $N_c \leq N/C$ clients for cluster $c$
%\ENDFOR
%\begin{center}
\STATE $F_{i}(0,1)$ = $\sum_{j=1}^N W_{j}P_{i}(0,1)$ 
\STATE  $L_{i}$ = $max(F_{i}(0),F_{i}(1))$
\STATEx
%\end{center}
\ENDFOR
\STATE Return $L_{i}$
\ENDFOR
\end{algorithmic}
\end{algorithm}
%%%%%%%%%%%%%%%%%%%%%%%%%%%%%%%%%%

In this work, three different optimization methods namely Particle Swarm Optimization (PSO), NelderMead, and Powell's method are employed to choose proper weights for the classification models. Moreover, for weight selection/optimization, a separate validation set is used for the computation of an accumulative classification error in the fitness function of each method. We note that the same fitness function, as shown in Equation \ref{fitness_function}, is used for all three methods employed in this work. In the equation, $error$ and $A_{acc}$ represent the classification error and accumulative accuracy computed on the validation set, respectively. 
%%%%%%%%%%%%%%%%%%%%%%%%%%%%%%%%%%%%%%%%%%%%%%
\begin{equation}
error = 1-A_{acc}
	\label{fitness_function}
\end{equation}
%%%%%%%%%%%%%%%%%%%%%%%%%%%%%%%%%%%%%%%%%
%The accumulative accuracy is computed using Equation \ref{equ:accuracy}, where $x_{1}, x_{2}, x_{3}, ... x_{n} $ are the variables (weight values) to be optimized while $p_{1}, p_{2}, p_{3}, ... p_{n} $ \textit{$p_{n}$} represents the probabilities obtained through the individual models on the validation set. 
%%%%%%%%%%%%%%%%%%%%%%%%%%%%%%%%%%%%%%%%%
%\begin{equation}
%%\small
%%\centering
%A_{acc} = x(1)*p_{1}+x(2)*p_{2}+... +x(n)*p_{n} 
%\label{equ:accuracy}
%\end{equation}
%%%%%%%%%%%%%%%%%%%%%%%%%%%%%%%%%%%%%%%%
Our choice for the weight selection/optimization is mainly based on the proven performance of the methods in similar applications \cite{ahmad2022social}. Some key aspects of these methods are provided below.

\begin{itemize}

\item \textbf{Particle Swarm Optimization-based Fusion} 
PSO \cite{kennedy1995particle} is one of the widely explored optimization methods and has been employed in a diversified set of applications for different tasks. For instance, it has been used for hyper-parameter optimization of Neural Networks (NNs) \cite{wang2019cpso} and Federated Learning (FL) \cite{qolomany2020particle}. The literature also reports its applicability in fusion to select/optimize weights to be assigned to the individual models/classifiers \cite{ahmad2022social}. The algorithm works in several steps starting with a random population of candidate solutions (i.e., different weight combinations). The algorithm then strives to iteratively improve the solution by evaluating the candidate solutions against the given criteria described in its fitness function. During the process, the algorithm keeps track of four parameters including personal and global-best, and the position and velocity of each particle. These parameters are updated at each iteration accordingly until a global minimum is obtained. Despite being very effective in different applications, there are several limitations of PSO. For instance, it is very slow and may fall into the local optimum in high dimensional space. However, in our case, we have a very low-dimensional problem.

%Our first merit-based fusion method is based on PSO \cite{kennedy1995particle}. PSO is heuristic in nature, which means that the solution (weights combination in our application) obtained with PSO is not globally optimal. However, the literature shows that the solution obtained with PSO is usually near the global optimal. The working mechanism of PSO is motivated by a group of fish or a flock of birds moving together, where the individual members of the group help each other by sharing their discoveries with the rest of the group leading to the best hunt for the entire group. In PSO, a similar approach is followed in three different steps. Firstly, an evaluation of each of the candidate solutions is carried out based on fitness criteria. Secondly, the personal best and global best values are updated. Finally, the position and velocity of each particle in the swarm are updated accordingly. %In this work, our fitness function is based on an accumulative classification error computed on a validation set using equation \ref{fitness_function}.

  \item \textbf{NelderMead Method-based Fusion}
Nelder Mead is also one of the oldest search/optimization algorithms. The name of the method is originated from the names of the researchers (Nelder and Mead) who introduced the algorithm in 1965. Similar to PSO, Nelder Mead has also been widely utilized in a diversified list of applications, and it is available in several ML and statistical analysis libraries \cite{ahmad2022social}. The method is suitable for n-dimensional problems (i.e., $n \geq 0$). Similar to PSO, the method can find the max/min of an objective function, which is based on the accumulative classification error on the validation, resulting in a combination of weights with better classification results. %The method is used to find the combination of weights to be assigned to the models that result in minimum classification error. 
Several implementations of the method are available in different libraries, however, we used SciPy\footnote{https://scipy.org/} implementation. 

\item \textbf{Powell's Method-based Fusion}
Powell's method is also one of the widely explored search and optimization methods. The method is based on the censored maximum-gradients technique. The method is very simple and can be employed in different applications. Similar to PSO, the method solves a problem iteratively by starting with an initial guess (i.e., a combination of weights) and moving towards a minimum of a function. During the process, the algorithm searches and calculates the distance in a good direction. The method also has some limitations, for instance, its assumption of the step size in the direction. In this work, we used SciPy\footnote{https://scipy.org/} implementation of the method, and our fitness function is defined in equation \ref{fitness_function}.
 
\end{itemize}
%\section{Evaluation Metrics}
%\label{sec:evaluation_metrics}

\section{Dataset and Experimental Setup}
\label{sec:experiments_setup}

\subsection{Dataset}
The proposed solutions are evaluated on a benchmark dataset released for PAN-21 authorship analysis shared task \cite{zangerle_eva_2021_4589145}. The dataset covers a total of 16,000 documents taken from the StackExchange network of Q\&A sites. For topical consistency, all the documents are taken from sites related to technology, such as code reviews, Computer Science, Data Science, etc. To ensure the quality of the data, several measures were taken by the task organizers. For instance, the text is cleaned by removing URLs, code snippets/screenshots, bullet lists, and documents having modified and altered text. Moreover, the text is organized into paragraphs, and to ensure sufficient text in each sample, paragraphs having less than  100 characters are removed from the documents. Moreover, the dataset contains an equal number of documents per author leading to an equal number of documents written by single, two, three, and four authors. This leads to a data imbalance problem in the first task where 25\% documents are single-authored and 75\% are multi-authored documents. 

We note that the same dataset, which is provided in separate training, validation, and test sets, is used for all the tasks. The training set contains 11,200 (70\%) while the test and validation sets are composed of 2,400 (15\% each).

\subsection{Experimental Setup}
The main objective of the work is to explore different aspects of stylometry/style change detection. Firstly, we aim to analyze the difficulty level of different tasks involved in stylometry/style change detection by evaluating the performance of the proposed methods in three different tasks. Secondly, we aim to analyze the impact of pre-processing/data cleaning on the performances of the proposed methods in all the tasks tackled in this work. We also want to analyze the performance of different NLP models and fusion methods to jointly employ these models for text classification. 

To achieve these goals, we conducted the following experiments using different experimental setups.   
\begin{itemize}
    \item We conducted experiments on three different tasks associated with stylometry/style change detection, each having a different difficulty level and challenges as discussed earlier.
    \item We conducted experiments on both clean and raw (un-clean) datasets. In the cleaned dataset, we mainly used some basic pre-processing techniques to clean the data. In our previous work \cite{zamir2023document}, we conducted an extensive data cleaning by removing stop words, URLs, short words, separate alphanumeric characters, and expansion of contractions. These techniques are generally used in different NLP applications to improve the performance of text classification algorithms. However, in this work, we conducted an extensive experiment to analyze which of the special characters have more impact on the performance of the models in the application.
    \item We conducted experiments to evaluate the performances of the individual models and several fusion methods including naive averaging and multiple merit-based late fusion methods. 
\end{itemize}

\section{Experimental Results}
\label{sec:results}
This section provides the experimental results of the proposed methods on all the tasks. We also provide comparisons against the existing solutions for each task.

\subsection{Single vs. Multiple Authors Classification}
Table \ref{tab:task1_results} provides the experimental results on both clean and raw (i.e., uncleaned) datasets in terms of F1-score. Moreover, the table reports the results of individual models as well as the fusion methods. As can be seen in the table, the performances on the raw data (i.e., the uncleaned dataset) are higher than the cleaned dataset. The maximum F1-scores obtained on the clean and raw datasets are 0.81 and 0.84, respectively, with a gap of around 3\%. This gap is lower than what we obtained in our initial work \cite{zamir2023document}. As discussed earlier, in this work, we expanded the scope of the experiments and explore the impact of different types of characters in the data-cleaning phase. Previously \cite{zamir2023document}, we cleaned the data by removing several special characters, such as usernames, URLs, emojis, stop words, and short words. Moreover, contractions are also expanded to their original form. %Generally, in NLP applications, the removal of these special characters improves the performances of NLP algorithms. 
However, in this work we observed that the removal of stop and short words has an adverse impact on the performance of the models and thus are kept in the data. The performances of the proposed methods improved significantly by retaining the stop and short words in the data. This improvement in the results indicates that these special characters help in differentiating authors' styles and thus lead to improved results in author style detection. 

As far as the evaluation of the individual models and fusion methods is concerned, overall better results are obtained with both versions of RoBERTa (i.e. RoBERTa-base and XML-RoBERTa) while comparable results are obtained with the base BERT and its variants. However, significant improvements in performance are observed when these models are jointly employed in fusion. The improvement in the performance over the best-performing individual model on the cleaned dataset is around 2\%, which is much higher compared to the gap on the raw dataset. Moreover, the late fusion methods used in this work can be broadly divided into naive (i.e., simple averaging) and merit-based fusion where weights of the models are obtained through the optimization methods described earlier. Overall, better results are obtained with the merit-based fusion with an improvement of around 2\% over the simple averaging. 
\begin{table}[]
\caption{Experimental results of the proposed methods on both datasets.} 
\label{tab:task1_results}
\begin{tabular}{|c|c|c|}
\hline
\textbf{Method/Model} & \textbf{Clean Dataset} & \textbf{Raw Dataset} \\ \hline
BERT &	0.79 & 0.7946 \\ \hline
Albert	& 0.80 & 0.7919 \\ \hline
 Distillbert & 0.76 & 0.7596 \\ \hline
 Roberta-base & 0.8344 & .80 \\ \hline
 XLM-Roberta	& 0.79 & 0.83 \\ \hline
Simple Fusion	& 0.770 & 0.8272 \\ \hline
 PSO-based Fusion & 0.790 & 0.8486 \\ \hline
Nelder-Mead Method-based Fusion & 0.810 & 0.8477 \\ \hline
Powell Method-based Fusion	& 0.811 & 0.8485 \\ \hline
\end{tabular}
\end{table}
%%%%%%%%%%%%%%%%%%%%%%%%%%%%%%%%%%%%%%%%%%%%%%%%%%%

%%%%%%%%Comparison Against existing methods %%%%%%%%%%%%
\begin{table}[]
\caption{Comparison against existing solutions on task 1.}
\label{tab:task1_comparison}
\begin{tabular}{|c|c|}
\hline
\textbf{Method} & \textbf{F1-Score} \\ \hline
 Zhang \cite{zhangstyle} & 0.753 \\ \hline
 Strom \cite{strom2021multi} & 0.795 \\ \hline
 Singh et al. \cite{singh2021writing} & 0.634 \\ \hline
  Deibel et al. \cite{deibel2021style} & 0.621 \\ \hline
   Nath \cite{nath2021style} & 0.704 \\ \hline
 Nelder-mead Method-based Fusion (This work) & 0.847 \\ \hline
PSO-based Fusion (This work) & 0.848 \\ \hline
Powell Method-based Fusion (This work) & 0.848 \\ \hline
\end{tabular}
\end{table}
%%%%%%%%%%%%%%%%%%%%%%%%%%%%%%%%%%%%%%%%%%%%%
Table \ref{tab:task1_comparison} provides the comparison of the proposed fusion methods against the existing solutions on task 1. We note that only our best-performing methods on the raw data are considered for comparison against the existing solutions. As can be seen, the best-performing method obtained an improvement of 5.3\% over the best existing solution in the literature. Some potential causes of the improvement include the fusion, choice of NLP models, and retaining of the special characters in the text. 

\subsection{Style Change Basic}
Table \ref{tab:task2_results} provides the experimental results on task 2 including the individual models and fusion methods' results on both clean and uncleaned datasets. A similar trend has been observed on task 2, where better results are obtained on raw data compared to the cleaned dataset. However, the difference in the performance of the best-performing methods on both datasets is higher (i.e., around 5\%) on task 2, clearly indicating the importance of the special characters in the application. 

As far as the performances of the individual models are concerned, there is no clear winner on both cleaned and raw datasets. However, overall XML-RoBERTa and ALBERT have slightly better results compared to the other models on both datasets. We also observed some variations in the performances of the models on clean and raw datasets. For instance, DistilBERT obtained comparable results on the raw dataset, however, it obtained a significantly lower F1 score compared to the other models on the cleaned data. Moreover, similar to task 1, the fusion methods also improved the results compared to the individual models. However, the improvement over the best-performing model is only around 2\%. One of the reasons for the less improvement in the result could be the adverse impact of the low-performing model (i.e., DistilBERT). 

%%%%%%%%%%%Individual Model Results %%%%%%%%%%%%%%%%%
\begin{table}[]
\caption{Experimental results of the proposed methods on both datasets for task 2.} 
\label{tab:task2_results}
\begin{tabular}{|c|c|c|}
\hline
\textbf{Method/Model} & \textbf{Clean Dataset} & \textbf{Raw Dataset} \\ \hline
BERT &	0.70528 & 0.7423 \\ \hline
Albert	& 0.72224 & 0.7719 \\ \hline
 Distillbert & 0.688 & 0.7245 \\ \hline
 Roberta-base & 0.7143 & 0.7618 \\ \hline
 XLM-Roberta	& 0.71294 & 0.7623 \\ \hline
Simple Fusion	& 0.7368 & 0.7822 \\ \hline
 PSO-based Fusion & 0.7117 & 0.7753 \\ \hline
Nelder Mead	method-based Fusion & 0.7117 & 0.7753 \\ \hline
Powell method-based Fusion	& 0.7224 & 0.7473 \\ \hline
\end{tabular}
\end{table}
%%%%%%%%%%%%%%%%%%%%%%%%%%%%%%%%%%%%%%%%%%%%%%%%%%%
We also provide comparisons against the existing solutions for task 2. As shown in Table \ref{tab:task2_comparison}, overall, most of our fusion methods outperformed the best-performing existing solution. Our best-performing method on the task (i.e., simple fusion) has an improvement of 3\% over the state-of-the-art on task 2. Some potential causes of the improvement include the fusion and retaining of the special characters in the text.

%%%%%%%%Comparison Against existing methods %%%%%%%%%%%%
\begin{table}[]
\caption{Comparison against existing solutions on task 2.}
\label{tab:task2_comparison}
\begin{tabular}{|c|c|}
\hline
\textbf{Method} & \textbf{F1-Score} \\ \hline
 Zhang \cite{zhangstyle} & 0.751 \\ \hline
 Strom \cite{strom2021multi} & 0.707 \\ \hline
 Singh et al. \cite{singh2021writing} & 0.657 \\ \hline
  Deibel et al. \cite{deibel2021style} & 0.669 \\ \hline
   Nath \cite{nath2021style} & 0.647 \\ \hline
    Simple Fusion (This work) & 0.782 \\ \hline
 Nelder-mead Method-based Fusion (This work) & 0.775 \\ \hline
PSO-based Fusion (This work) & 0.775 \\ \hline
Powell Method-based Fusion (This work) & 0.747 \\ \hline
\end{tabular}
\end{table}
%%%%%%%%%%%%%%%%%%%%%%%%%%%%%%%%%%%%%%%%%%%%%

\subsection{Style Change Real-World}
Table \ref{tab:task3_results} provides our results on the third task. As can be observed in the table, in contrast to the first two tasks, comparable results are obtained on both clean and raw datasets. The performances of the proposed methods on the cleaned dataset are even better in some cases. For instance, the performances of the PSO and Nelder Mead method-based fusion on the cleaned dataset are better than all the methods on the raw dataset. Moreover, in the individual models, the BERT-base model obtained the highest score while PSO and Nelder Mead-based fusion methods are the best-performing fusion methods on the task.
%%%%%%%%%%%Individual Model Results %%%%%%%%%%%%%%%%%
\begin{table}[]
\caption{Experimental results of the proposed methods on both datasets for task 3.} 
\label{tab:task3_results}
\begin{tabular}{|c|c|c|}
\hline
\textbf{Method/Model} & \textbf{Clean Dataset} & \textbf{Raw Dataset} \\ \hline
BERT &	0.5309 & 0.5244 \\ \hline
Albert	& 0.4998 & 0.494 \\ \hline
 Distillbert & 0.5261 & 0.5213 \\ \hline
 Roberta-base & 0.48 & 0.47 \\ \hline
 XLM-Roberta	& 0.5281 & 0.5159 \\ \hline
Simple Fusion	& 0.5087 & 0.5164 \\ \hline
 PSO-based Fusion 	& 0.5535 & 0.5404 \\ \hline
Nelder-mead Method-based Fusion	& 0.5535 & 0.5404 \\ \hline
Powell Method-based Fusion	& 0.5453 & 0.5323 \\ \hline
\end{tabular}
\end{table}
%%%%%%%%%%%%%%%%%%%%%%%%%%%%%%%%%%%%%%%%%%%%%%%%%%%

We also provide a comparison against the existing methods for task 3. As can be observed in Table \ref{tab:task3_comparison}, our best-performing methods PSO and Nelder Mead-based fusion methods have an improvement of around 5\% over the best-performing existing solutions.

%%%%%%%%Comparison Against existing methods %%%%%%%%%%%%
\begin{table}[]
\caption{Comparison against existing solutions on task 3.}
\label{tab:task3_comparison}
\begin{tabular}{|c|c|}
\hline
\textbf{Method} & \textbf{F1-Score} \\ \hline
 Zhang \cite{zhangstyle} & 0.501 \\ \hline
 Strom \cite{strom2021multi} & 0.424 \\ \hline
 Singh et al. \cite{singh2021writing} & 0.432 \\ \hline
  Deibel et al. \cite{deibel2021style} & 0.263 \\ \hline
   %Nath et al. \cite{nath2021style} & 0.647 \\ \hline
    %Simple Fusion (This work) & 0.782 \\ \hline
 Nelder-mead Method-based Fusion (This work) & 0.5404 \\ \hline
PSO-based Fusion (raw-dataset, this work) & 0.5404 \\ \hline
PSO-based Fusion(clean-dataset, this work) & 0.5535 \\ \hline
Nelder-mead Method-based Fusion	(clean-dataset, this work) & 0.5535 \\ \hline
\end{tabular}
\end{table}
%%%%%%%%%%%%%%%%%%%%%%%%%%%%%%%%%%%%%%%%%%%%%

\subsection{Lessons Learned}
Some key lessons learned from this work can be summarized as follow.

\begin{itemize}
    \item As indicated by the experimental results, the complexity and the performances of the NLP models vary on the different tasks involved in SCD. The classification of single and multi-authored documents is more straightforward compared to the identification of text segments where the author switches. Similarly, the performance of the proposed methods is significantly lower on the third task involving the identification of multiple locations where the author switches.
    \item The individual models complement each other when jointly combined in the fusion, which leads to an improvement in the performance.
    \item The special characters and other words, such as stop and short words, which are generally removed in different NLP applications, play a vital role in the SCD. More, specifically, we observed the removal of stop and short words significantly reduced the performance of the proposed methods.
\end{itemize}

\section{Conclusions}
\label{sec:conclusion}
In this paper, we tackled three different tasks of stylometry/style change detection by proposing a merit-based fusion framework to jointly employ multiple NLP algorithms from text classification. In all three tasks, the performance of several transformer-based models and late fusion methods. We also analyzed the impact of pre-processing/cleaning on the performances of the proposed solutions in the tasks. The performance of the proposed methods on all the tasks demonstrates the potential of NLP algorithms in automatic SCD, which can lead to several interesting real-world applications. 

In the future, we aim to further explore the change detection by tasks by exploring other relevant subtasks in standard and informal (e.g., social media) texts. We will also explore the potential of these methods in differentiating text written by human authors and chatbots or text generators, such as ChatGPT.  

\section{Funding and/or Conflicts of interests/Competing interests}
The authors declare no conflict of interest. 
\section{Data Availability}
Data sharing is not applicable to this article as no datasets were generated or analyzed during the current study. The experiments are carried out on a publicly available dataset, which is cited in the paper.
%\begin{acknowledgements}
%The authors declare no conflict of interest
%\end{acknowledgements}

\bibliographystyle{spmpsci}      % mathematics and physical sciences

\bibliography{sigproc}

\end{document}